# PATHOLOGY FOUNDATION MODELS FOR TRAINING-FREE CASE RETRIEVAL IN SALIVARY GLAND TUMORS

**Maximilian J. Roesel** [1,2,*], **Niklas Kiermeyer** [1,2], **Philipp Keyl** [1,3,4], **Frederick Klauschen** [1,2,3,4]

[1] Institute of Pathology, Ludwig-Maximilians-Universität München

[2] German Cancer Consortium (DKTK), German Cancer Research Center (DKFZ), Munich partner site

[3] Institute of Pathology, Charité-Universitätsmedizin Berlin

[4] BIFOLD-Berlin Institute for the Foundations of Learning and Data

[*] Corresponding author: max.roesel@lmu.de

## ABSTRACT

In pathology, pattern recognition relies on a frame of reference built through extensive training and experience, which may still fall short when encountering rare cancer subtypes. Archives of previously diagnosed tumors often contain relevant precedents, yet identifying the most relevant cases remains challenging. Here we explore whether content-based retrieval with current pathology foundation models facilitates rare cancer diagnostics without any task-specific training. Pathologists can query a digitized archive using a region of interest (ROI) from a whole slide image (WSI) and retrieve morphologically similar cases, along with their ICD-O codes and corresponding WSIs. The ROI can be selected manually or automatically using a gated attention-based multiple instance learning model. We evaluate our approach by leave-one-case-out on an imbalanced dataset of 137 salivary gland tumor cases spanning 12 ICD-O morphology categories, comparing four pathology foundation models (Atlas2, Atlas2-B, Atlas2-S, and UNI2h). Top-1/3/5 retrieval was 62.5%/76.7%/81.3% for Atlas2, 50.2/66.6/72.9% for UNI2h, 52.3%/71.5%/77.6% for Atlas2-B and 54.4/65.2/73.7% for Atlas2-S. Macro-averaged F1 ranged from 69.0% for the smallest model Atlas2-S to 76.6% for the largest model Atlas2 across the five classes with at least ten cases. Our modified Chamfer similarity separated correct from incorrect matches among the Top-5 more reliably for Atlas2 (AUC 0.754) than for Atlas2-B (0.679, p = 0.020) and Atlas2-S (0.653, p = 0.003), but not UNI2h (0.697, p = 0.196). Requiring no additional training or manual annotation, this approach makes departmental archives searchable with pathology foundation models, providing a low-barrier route to case-based differential diagnosis and extending the individual frame of reference to the collective experience of the department.

## Introduction

Even as increasingly detailed methods of tissue characterization move toward clinical application, morphology remains the foundation of microscopic pathology (1). Mastering its interpretation requires years of training, building an internal frame of reference against which unseen cases are compared (2). That frame of reference, however, is bounded by the cases an individual has encountered. Each rare entity is, by its nature, seldom seen, yet rare entities collectively make up a substantial part of the diagnostic spectrum. And while any individual pathologist encounters a given entity only occasionally, a department accumulates such cases over years. Once digitized as whole slide images (WSIs), these institutional archives can become a valuable external frame of reference, yet searching through these archives is challenging due to the diverse morphology and size of WSIs (3).

Digitization has made histomorphology accessible to artificial intelligence (AI), which is advancing from research toward routine diagnostics (4). Neural networks can learn semantically meaningful representations of tissue directly from data, an approach known as deep learning (5, 6). Rather than learning a single task, foundation models (FMs) use self-supervised pre-training on large corpora of slides to learn general-purpose representations that can be used across downstream tasks (7, 8). While training classifiers is a common way to apply such representations, they require enough labelled cases per class to learn a decision boundary and return a label without morphological evidence behind it (9). Content-based retrieval avoids this by returning cases rather than labels (10-12). We built a search engine based on FM representations, letting pathologists query an institutional archive and explore whether current pathology FMs make this viable for rare entities without any task-specific training.

We evaluate our approach on salivary gland tumors as a proof-of-concept, a rare and morphologically challenging entity with an incidence of 5.2 per 1,000,000 person-years in Germany (13) and the WHO recognizing 21 malignant types (14), in addition to secondary tumors (15). While some entities cannot be reliably separated by their morphology (16) and new molecular findings continue to carve out new entities (17), initial histological assessment can guide the selection of ancillary testing. An archive of 137 cases spanning 12 ICD-O morphology categories was embedded with four pathology FMs and retrieval was evaluated using queries selected by a pathologist and by a gated attention-based multiple instance learning (ABMIL) model.

## Methods

### Data

We digitized 137 salivary gland tumor cases, totaling 974 hematoxylin and eosin (H&E) WSIs, for our test cohort. For training our ABMIL model, we digitized an additional 15 non-tumor cases, totaling 123 H&E WSIs. All scans were acquired at a native magnification of 40× with a resolution of 0.25 µm/pixel (microns per pixel, (MPP)) using Leica Aperio GT 450 DX scanners. ICD-O codes were sourced through the Zentrum für Krebsregisterdaten at the Robert Koch Institute and through our patient information system.

### Feature extraction

Every WSI was screened for artifacts using GrandQC (18). Tissue was tiled into nonoverlapping 512×512-pixel patches at 20× magnification covering 256×256 µm of tissue (0.5 MPP) using PathoPatcher (19). The patch size was chosen to balance morphological detail against index size (20). Patches had to contain at least 50% artifact-free tissue. No stain normalization was applied.

Patch-level features were extracted with four pathology FMs, all Vision Transformers (ViT) pretrained with the DINOv2 self-supervised framework and used here as frozen encoders (21). UNI2h, released under the CC-BY-NC-ND 4.0 license, is an open-source model with a custom ViT-H architecture with 681M parameters (24 blocks, 24 heads, SwiGLU feed-forward layers, eight register tokens), pretrained on over 350,000 H&E and immunohistochemistry (IHC) slides from Mass General Brigham (7). Atlas2 along with its distilled versions Atlas2-B and Atlas2-S was used under a research license. Atlas2 is the largest model with ~2 billion parameters pretrained with an adapted DINOv2 recipe on 5,5 million WSI spanning H&E, special stains, and IHC from institutions across Germany and the US (8).

To extract the feature vectors each 512×512-pixel patch is resized to a 224×224-pixel RGB image, partitioned into non-overlapping 14×14 pixel tiles, resulting in a sequence of 256 tile tokens, and prepended with a learnable class token (CLS) alongside register tokens (22). This token sequence is passed through the frozen ViT backbone. The tile and register tokens are discarded, and the final layer-normalized CLS embedding $z \in \mathbb{R}^d$ is taken as the patch descriptor. The dimensions ($d$) differ by the FM with $d$ = 384 for Atlas2-S, $d$ = 768 for Atlas2-B, $d$ = 1536 for UNI2h, and $d$ = 2048 for Atlas2.

ABMIL model

To train the gated ABMIL, patch-level features were extracted from tumor positive and negative slides with our standard specifications. Our training data contained 123 negative bags consisting of 294,448 patches and 137 positive bags consisting of 3,158,361 patches. A bag is defined as a matrix $H \in \mathbb{R}^{(n\times d)}$. For negative bags, $n$ is the number of patches for that slide. For positive bags, $n$ is the number of patches for that case sourced from a varying number of WSIs.

To compute patch-level attention weights we use the gated attention formulation (23). Two independent linear projections are applied to $H$: first, a tanh branch $V \in \mathbb{R}^{(d\times L)}$, capturing feature relevance; second, a sigmoid branch $U \in \mathbb{R}^{(d\times L)}$, acting as a gating mechanism to suppress irrelevant patches. These are combined by element-wise multiplication, projected to a scalar attention score per patch via a weight vector, and normalized with softmax across all patches in the bag to yield attention weights $a_i$. The resulting attention-weighted average of patch embeddings forms the bag-level representation $M$, which is passed through a single linear classifier $w$ to produce the bag-level prediction.

The same classifier applied to an individual patch embedding $h_i$ yields a per-patch tumor score $w\cdot h_i$, and $a_i$ $(w\cdot h_i)$ is that patch's contribution to the bag-level decision. Under bag-level supervision alone, attention and this per-patch score can decouple, so in addition to the main bag-level binary cross-entropy (BCE) classification loss we optimize two auxiliary losses. The first is an instance-alignment loss: patches are ranked by the detached per-patch score $w\cdot h$, and the loss is the negative log of the total attention mass placed on the top-k of them, pushing attention onto the patches the classifier itself reads as tumor while leaving the classifier untouched. This inverts CLAM's instance-level supervision, which ranks patches by attention to generate pseudo-labels for a separate instance classifier (24). The second exploits the fact that every patch of a tumor-free case is a confirmed negative instance. For each positive bag, the top-k patches by $w\cdot h$ are compared against m sampled certain negatives: those exceeding the highest-scoring negative are labeled positive, and a BCE loss is applied against the sampled negatives as ground truth. If no top-k patch exceeds that bar, the loss falls back to negatives-only. Unlike the alignment term, this loss updates the classifier weights w directly. For both terms, k follows max(8, f·n), where f is a tuned fraction that scales selection with bag size, combined with a fixed floor that prevents degenerate selection on small bags. Attention is detached for negative bags, as a tumor-free label provides no supervisory signal for attention placement. Mild class imbalance is addressed through a weighted random sampler; the positive class weight is set to 1.

The following hyperparameters were optimized jointly via Bayesian optimization: attention projection dimension, learning rate, weight decay, dropout rate, and early-stopping patience, along with the loss weights, top-k fraction, and number of sampled negatives for the two auxiliary losses. Validation-set loss was used as the optimization objective; Area under the receiver operating characteristic curve (AUC) was excluded, as it saturated on the small validation split before the model converged. The optimized configuration was then evaluated via five-fold cross-validation.

Once the final models were trained, attention weights and per-patch scores were computed across all tumor cases in a single forward pass. Because tumor is spatially contiguous whereas isolated high-scoring patches usually reflect noise, scores were averaged over each patch's 3×3 grid neighborhood before ranking. From these smoothed scores, we reconstructed the top spatially coherent 2×2 (1024×1024-pixel) and 4×4 (2048×2048-pixel) squares per case.

FAISS index

The database indices were built per FM using FAISS (25). Per-patch embeddings were first concatenated into a single matrix together with a parallel array of patch identifiers for metadata mapping. Each embedding was L2-normalized to unit length so that inner-product equals cosine-similarity.

We used an IVFPQ structure, which combines two complementary approximations. First, an inverted file index (IVF) partitions the embedding space into $\sqrt{N}$ Voronoi cells whose centroids are learned by k-means; at query time only the cells nearest a query are scanned, avoiding an exhaustive comparison against every vector. Second, product quantization (PQ) splits each vector's residual from its assigned centroid into $M$ contiguous sub-vectors. Each sub-vector is then replaced by its nearest centroid index from a learned 256-entry subspace codebook. The sub-vector length is held constant at 16 dimensions across all FMs so that quantization granularity remains comparable; $M$ is set accordingly yielding a fixed 64× compression ratio (24 for Atlas2-S, 48 for Atlas2-B, 96 for UNI2h, and 128 for Atlas2).

Both the IVF centroids and the PQ codebooks were trained on a random subsample of the corpus ($50\times\sqrt{N}$, >100,000) before the full set was added in batches. Vectors were inserted with sequential integer IDs equal to their row position, so a single metadata array maps each FAISS ID directly to its filename.

Retrieval

A query region is either sourced by our ABMIL model or by a pathologist. For a query, capture magnification and scanning specification (magnification and MPP) need to be specified. The scanning MPP defines the physical tissue area represented by one pixel at the original scan resolution; multiplying it by the ratio of scanning to capture magnification yields the effective tissue-per-pixel of the query image. This ensures that query and database patches represent tissue at the same physical scale prior to embedding.

The query region is divided into non-overlapping 512 × 512-pixel patches and embedded by a FM, yielding $n$ descriptors, each L2-normalized to unit length. These $n$ descriptors are pooled adaptively to avoid diluting distinct morphologies while still consolidating redundant ones. If the pairwise cosine similarity is above 0.8 the query is treated as homogeneous and all descriptors are averaged into a single re-normalized query vector. For a heterogeneous query, the descriptors are greedily clustered: proceeding through the descriptors in order, the first seeds a cluster and every remaining within 0.8 cosine similarity of that seed is merged into it, and the cluster mean is re-normalized to a query centroid; the pass repeats until all descriptors are assigned. The number of queries equals the number of clusters formed, between one and the descriptor count. The threshold, governing both the homogeneity gate and cluster merging, is tunable.

Each query vector is searched against the corresponding FAISS index. The closest clusters to the input image are found via k-means. The exact similarity within this cluster is then calculated via inner product similarity. The number of probed Voronoi cells is dynamically set to min(max(200, ⌊clusters/5⌋), 1500), balancing recall against search latency. A patch-level adaptable candidate pool is retrieved per query (default = 150,000).

Retrieved patches are grouped by their source case. Each candidate is ranked by a sparse Chamfer score computed only over the patches retrieved. Chamfer distance scores two sets by summing each element's distance to its nearest neighbor in the other set, in both directions. Because patch embeddings are L2-normalized, nearest-neighbor matching under Euclidean distance and under cosine similarity is equivalent. The two principal terms are asymmetric as the numbers of patches per query and retrieved case vary. Thus, we use a normalized forward and backward term. The forward term averages, over query vectors, each query vector's cosine similarity to its single best-matching retrieved patch, measuring whether the case covers the query. The backward term averages, over the retrieved patches, each patch's cosine similarity to its single best-matching query vector, measuring whether the query is supported across the retrieved patches rather than by one coincidental match. This factor gets adjusted by a saturating term $(1 - e^{-n/5})$ favoring more retrieved patches and saturating at around 15 patches. The final ranking score is the sum of the forward and the adjusted backward term.

<u>Evaluation</u>

All statistical tests used $\alpha = 0.05$, with Holm correction applied within each family of comparisons; the composition of each family is stated at the relevant test below.

We compared FMs via leave-one-out cross-validation across query-source model and index model. For 2×2 and 4×4 contiguous regions, we report macro-averaged recall over all classes at k = 1, 3, 5; a retrieval is positive if at least one case of the correct class appears among the k results. Model pairs were compared via paired t-test on per-case recall differences, Holm-corrected across the six pairs, within each depth and region scope (2×2, 4×4, pooled). We additionally evaluated 4×4 regions selected by a pathologist, reporting recall at k = 1, 3, 5 for all classes (compared with the same paired t-test) and precision, recall, and F1 at k = 1 for the five classes with ≥ 10 cases, tested as described below.

Within each index model, we compared Chamfer score AUC for correct vs. incorrect retrievals, pooling each query's top-5 with a case-level bootstrap (2,000 resamples of 137 queries, jointly across models). Precision, recall, and F1 at k = 1 used the same bootstrap. Between-model differences were tested on the resulting bootstrap difference distribution, Holm-corrected across six pairs per metric; each model's AUC was also tested against chance (0.5) via one-sided Wald test on bootstrap standard error, Holm-corrected across four models.

For the seven classes with ≥ 5 cases, chance was defined as the hypergeometric probability that k random draws without replacement yield at least one positive retrieval; each class's aggregate hit count was tested against this null with a one-sided exact binomial test. A class was called significant only if this held for all four indexing models, Holm-corrected across the 21 class-by-depth combinations.

## Results

To test whether current pathology FMs allow pathologists to query an institutional archive without task-specific training, we developed the following search engine (Fig. 1). Users can select a region of interest (ROI) in two ways: automatically, using our ABMIL model that identifies the most diagnostically relevant, spatially contiguous region; or manually, by selecting a rectangular ROI of arbitrary size from a WSI, allowing the pathologist to query tissue they consider diagnostically relevant (Fig. 1a).

This ROI is divided into non-overlapping patches (Fig. 1b), and each patch is converted by a FM into a numerical representation that captures its visual features (Fig. 1c). These representations let us measure how similar any two patches are to each other. If the patches within the region are highly similar to one another, the region is treated as homogeneous, and the representations are combined into a single representative query. If the patches are dissimilar, the region is treated as heterogeneous: the patches are grouped into subsets of visually similar tissue, and each subset contributes its own representative query (Fig. 1d). This step reduces redundancy from near-duplicate patches while still allowing heterogeneous tissue to be searched. A homogeneous region therefore generates one query, while a heterogeneous region generates several. Each query is then compared against the representations of previously diagnosed archive cases, extracted in advance using the same FM (Fig. 1e).

Matches are scored using our modified Chamfer similarity (Fig. 1f), and the pathologist is shown a ranked list of the most similar archive cases, each labeled with its ICD-O diagnosis. For any retrieved case, the individual patches driving the match can be inspected at their original location within the slide, letting the user verify that the retrieved tissue is genuinely comparable.

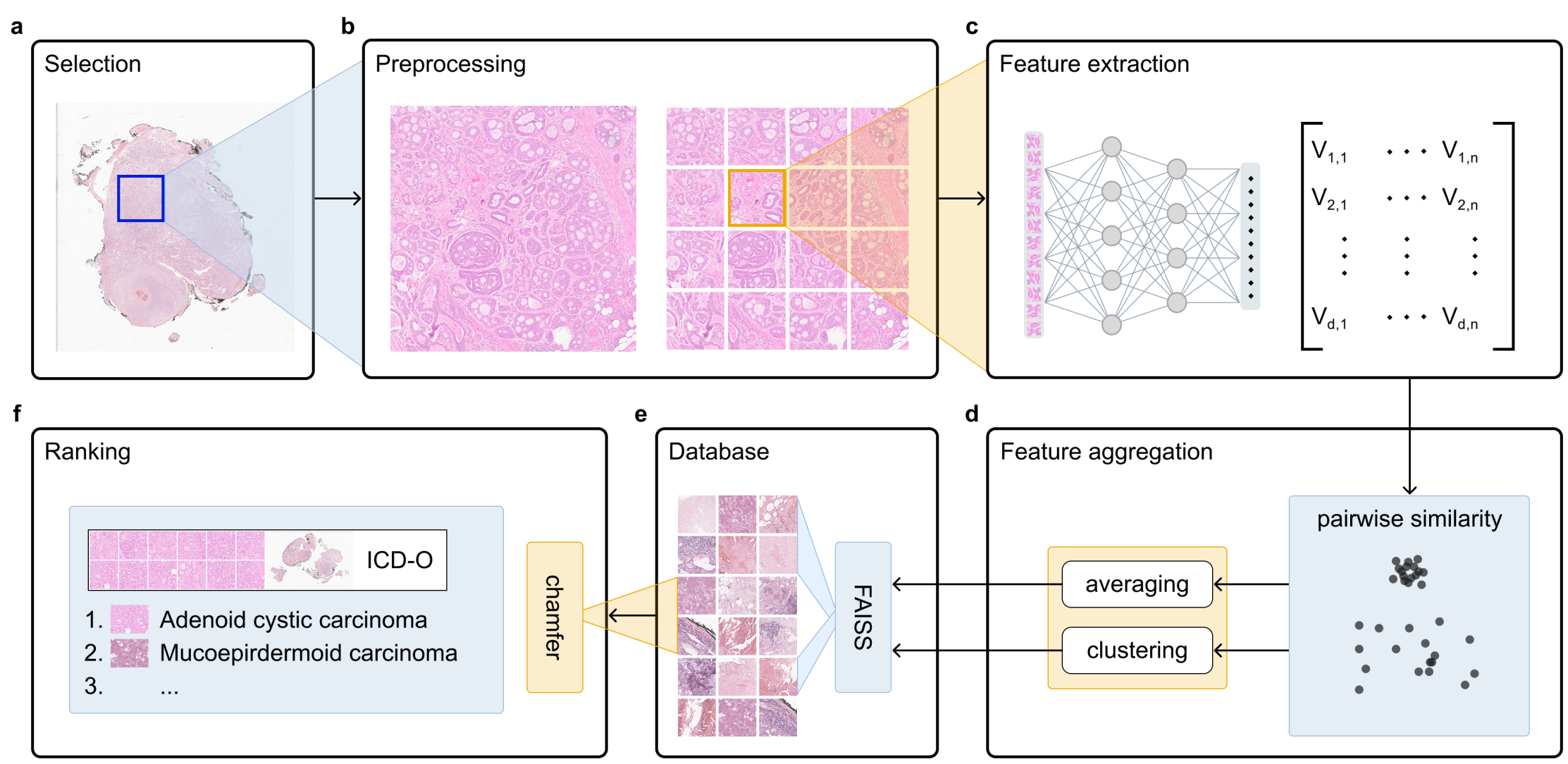

**Figure 1 – Overview of the search engine.** (a) Starting from a WSI, a region of interest can either be selected by a pathologist or automatically through attention-based multiple instance learning. (b) The region of interest is tiled into non-overlapping 512×512-pixel patches. (c) Each patch is passed through a foundation model, producing a matrix of patch-level embeddings. (d) Based on pairwise similarity among the query embeddings, feature vectors are either averaged or greedily clustered. (e) Query vector/s are searched against a FAISS index of pre-extracted patch embeddings. (f) Retrieved patches are scored by a modified Chamfer similarity, returning each match's ICD-O diagnosis alongside its WSI. Abbreviations: d = embedding dimension of the foundation model. n = number of patches after tiling. ICD-O = International Classification of Diseases for Oncology.

To evaluate our search engine, we used a dataset of 137 primary and secondary salivary gland tumors spanning 12 ICD-O morphologies (Fig. 2). The dataset is imbalanced, with 82% of cases concentrated in five morphologies (Fig. 2a). We excluded entities represented by only a single case but deliberately retained the three morphologies with only two cases for a both statistically and morphologically challenging task (Fig. 2b).

While this search engine requires no model training, we wanted to know how well automated sourcing through ABMIL performed for each FM. We therefore sourced the region with the highest average attention at both 2×2- and 4×4-patch sizes (Fig. 2c). Neither the FM used for sourcing, nor the query size produced a relevant difference in retrieval accuracy. Averaged over all 16 configurations, 4×4 queries tended to be slightly better than 2×2 at Top-1 (50.3% vs. 48.3%, $p = 0.110$) and were essentially identical at Top-3 (64.3% vs. 63.8%, $p = 0.691$) and Top-5 (71.5% vs. 71.2%, $p = 0.831$). As query size was therefore not a performance-driven decision, we selected 4×4 for the remaining analyses, as it corresponds better to the field of view a pathologist typically assesses.

The pathology FM used for retrieval did affect performance (Fig. 2d). Pooled across both query sizes, Atlas2 retrieved the correct morphology at Top-1 more often than Atlas2-S (53.5% vs. 47.4%, $p = 0.029$) and UNI2h (53.5% vs. 47.9%, $p = 0.029$) but not significantly more often than Atlas2-B (53.5% vs. 48.5%, $p = 0.074$). There was no significant difference at Top-3 or Top-5 between Atlas2 (64.5%, 74.7%), UNI2h (61.8%, 66.9%), Atlas2-B (64.9%, 73.1%) and Atlas2-S (65.0%, 70.7%).

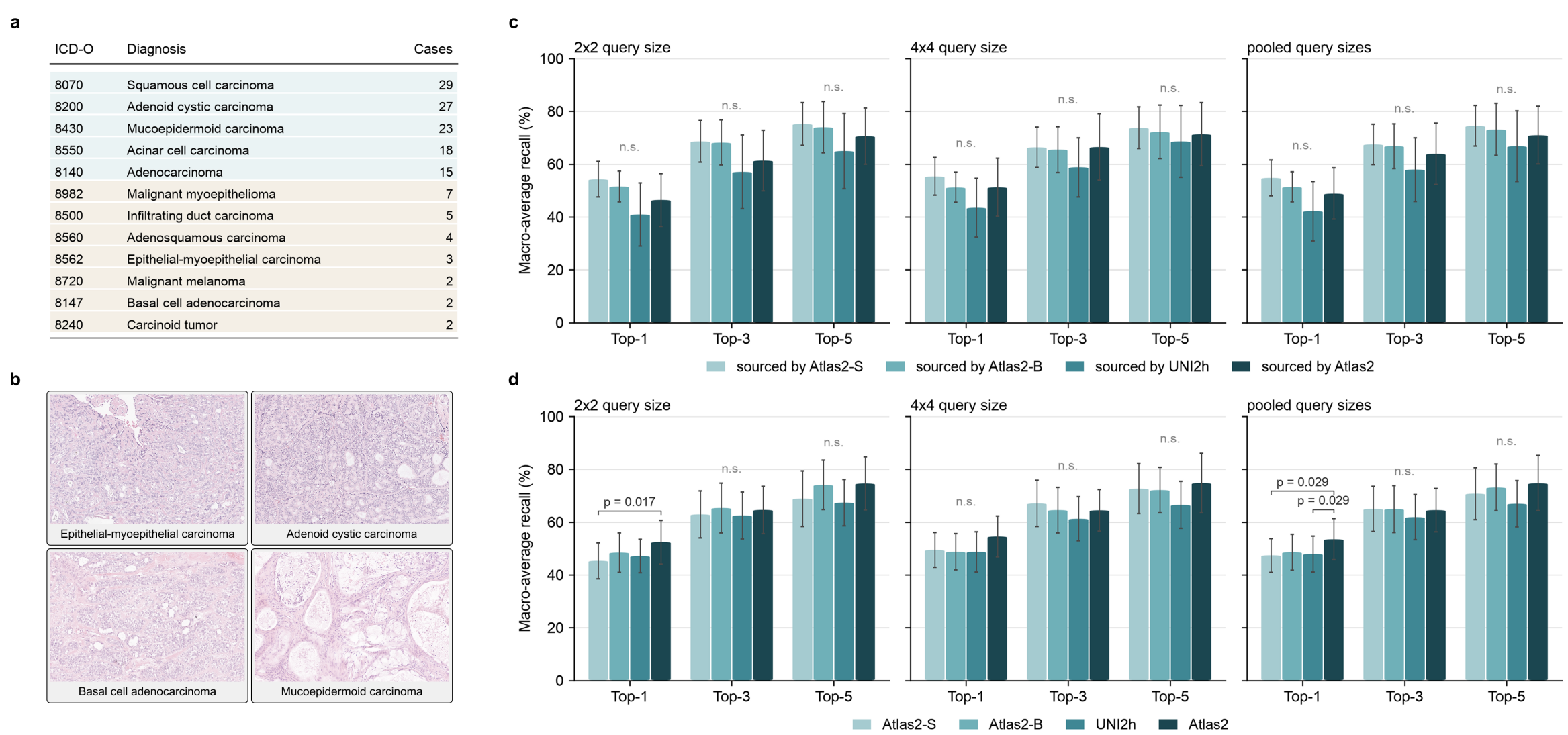

**Figure 2. Cohort composition and retrieval performance by sourcing and indexing model.** (a) Evaluation cohort, beige color indicates less than ten cases. (b) Representative examples of four morphology classes. (c) Macro-averaged recall by query size and by which model the input was sourced, retrieval averaged across all models. (d) Macro-averaged recall by query size and by which model was used for retrieval, input averaged across all models. Error bars are 95% confidence intervals for a stratified average. n.s. = not significant after Holm correction. 2×2 = 1024×1024-pixel query, 4×4 = 2048×2048-pixel query. ICD-O = International Classification of Diseases for Oncology.

Next, we examined per-class behavior on the same cohort, using ROIs selected by a pathologist (Fig. 3). Macro-averaged recall at Top-1/3/5 was 62.5%/76.7%/81.3% for Atlas2, 50.2%/66.6%/72.9% for UNI2h, 52.3%/71.5%/77.6% for Atlas2-B, and 54.4%/65.2%/73.7% for Atlas2-S, without any statistically significant difference between the models at any depth. Performance was not dependent on class size with malignant melanoma and carcinoid tumor, with only two cases each, retrieved correctly by every model at Top-1. Conversely, basal cell adenocarcinoma, also with only two cases, was not retrieved correctly by any model at Top-1 (Fig. 3a).

Restricted to classes with at least ten cases, precision and recall were balanced except for adenocarcinoma, which was under-retrieved rather than over-called. F1 was also where the models diverged most: squamous cell carcinoma was essentially model-independent, whereas adenocarcinoma ranged from 0.36 (Atlas2-S) to 0.62 (Atlas2) and mucoepidermoid carcinoma from 0.62 (UNI2h) to 0.78 (Atlas2). Across these five classes, macro-averaged F1 was 76.6% for Altas2, 71.8% for UNI2h, 74.0% for Atlas2-B and 69.0% for Atlas2-S without any statistically significant difference (Fig. 3b).

Four of the seven classes with at least five cases achieved above-chance Top-1 and Top-3 retrieval across all four models ($p < 0.05$), while adenocarcinoma, malignant myoepithelioma, and infiltrating duct carcinoma only tended to exceed chance (Fig. 3c).

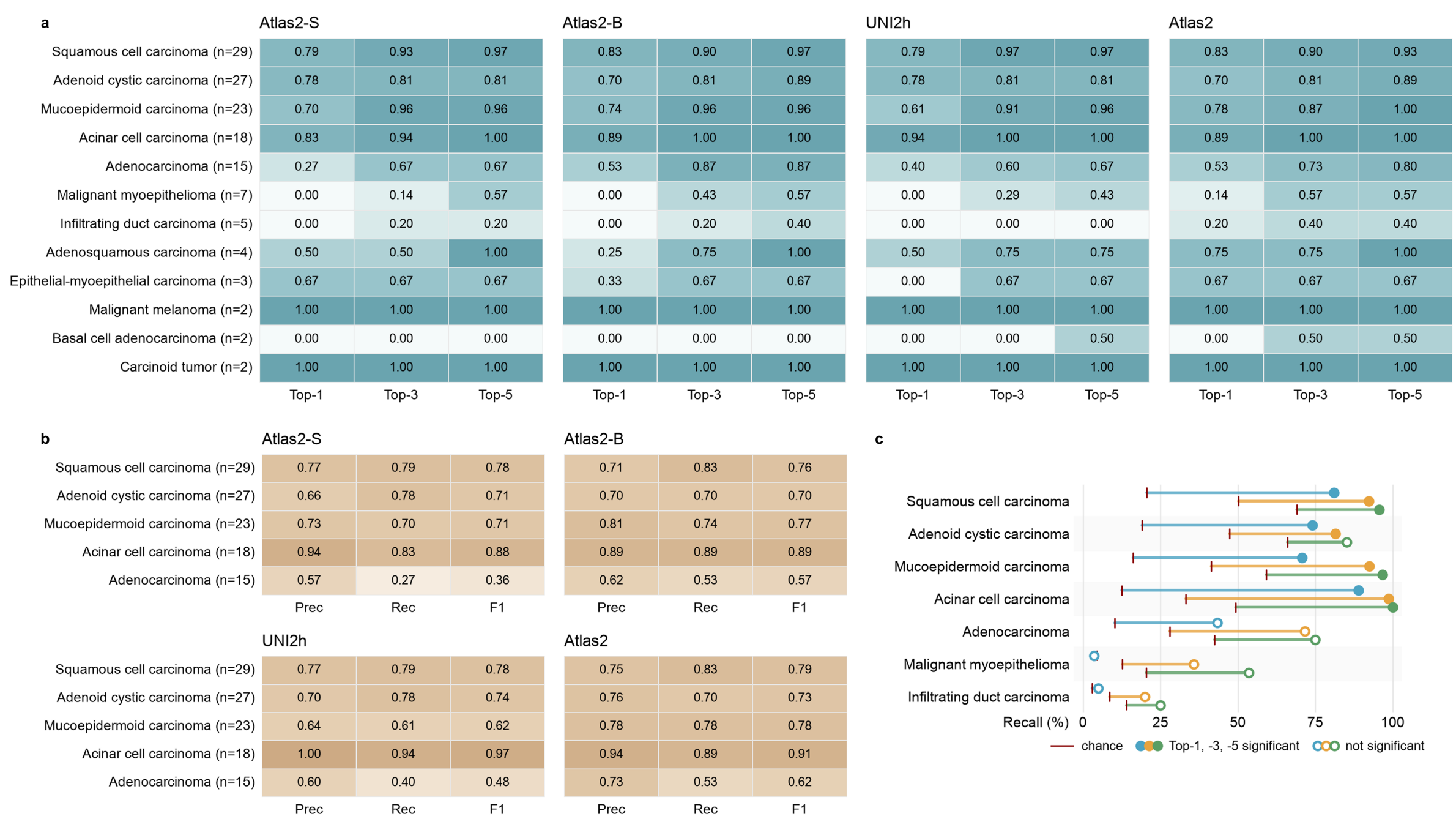

| Atlas2-S | Top-1 | Top-3 | Top-5 |
|---|---|---|---|
| Squamous cell carcinoma (n=29) | 0.79 | 0.93 | 0.97 |
| Adenoid cystic carcinoma (n=27) | 0.78 | 0.81 | 0.81 |
| Mucoepidermoid carcinoma (n=23) | 0.70 | 0.96 | 0.96 |
| Acinar cell carcinoma (n=18) | 0.83 | 0.94 | 1.00 |
| Adenocarcinoma (n=15) | 0.27 | 0.67 | 0.67 |
| Malignant myoepithelioma (n=7) | 0.00 | 0.14 | 0.57 |
| Infiltrating duct carcinoma (n=5) | 0.00 | 0.20 | 0.20 |
| Adenosquamous carcinoma (n=4) | 0.50 | 0.50 | 1.00 |
| Epithelial-myoepithelial carcinoma (n=3) | 0.67 | 0.67 | 0.67 |
| Malignant melanoma (n=2) | 1.00 | 1.00 | 1.00 |
| Basal cell adenocarcinoma (n=2) | 0.00 | 0.00 | 0.00 |
| Carcinoid tumor (n=2) | 1.00 | 1.00 | 1.00 |

| Atlas2-B | Top-1 | Top-3 | Top-5 |
|---|---|---|---|
| Squamous cell carcinoma (n=29) | 0.83 | 0.90 | 0.97 |
| Adenoid cystic carcinoma (n=27) | 0.70 | 0.81 | 0.89 |
| Mucoepidermoid carcinoma (n=23) | 0.74 | 0.96 | 0.96 |
| Acinar cell carcinoma (n=18) | 0.89 | 1.00 | 1.00 |
| Adenocarcinoma (n=15) | 0.53 | 0.87 | 0.87 |
| Malignant myoepithelioma (n=7) | 0.00 | 0.43 | 0.57 |
| Infiltrating duct carcinoma (n=5) | 0.00 | 0.20 | 0.40 |
| Adenosquamous carcinoma (n=4) | 0.25 | 0.75 | 1.00 |
| Epithelial-myoepithelial carcinoma (n=3) | 0.33 | 0.67 | 0.67 |
| Malignant melanoma (n=2) | 1.00 | 1.00 | 1.00 |
| Basal cell adenocarcinoma (n=2) | 0.00 | 0.00 | 0.00 |
| Carcinoid tumor (n=2) | 1.00 | 1.00 | 1.00 |

| UNI2h | Top-1 | Top-3 | Top-5 |
|---|---|---|---|
| Squamous cell carcinoma (n=29) | 0.79 | 0.97 | 0.97 |
| Adenoid cystic carcinoma (n=27) | 0.78 | 0.81 | 0.81 |
| Mucoepidermoid carcinoma (n=23) | 0.61 | 0.91 | 0.96 |
| Acinar cell carcinoma (n=18) | 0.94 | 1.00 | 1.00 |
| Adenocarcinoma (n=15) | 0.40 | 0.60 | 0.67 |
| Malignant myoepithelioma (n=7) | 0.00 | 0.29 | 0.43 |
| Infiltrating duct carcinoma (n=5) | 0.00 | 0.00 | 0.00 |
| Adenosquamous carcinoma (n=4) | 0.50 | 0.75 | 0.75 |
| Epithelial-myoepithelial carcinoma (n=3) | 0.00 | 0.67 | 0.67 |
| Malignant melanoma (n=2) | 1.00 | 1.00 | 1.00 |
| Basal cell adenocarcinoma (n=2) | 0.00 | 0.00 | 0.50 |
| Carcinoid tumor (n=2) | 1.00 | 1.00 | 1.00 |

| Atlas2 | Top-1 | Top-3 | Top-5 |
|---|---|---|---|
| Squamous cell carcinoma (n=29) | 0.83 | 0.90 | 0.93 |
| Adenoid cystic carcinoma (n=27) | 0.70 | 0.81 | 0.89 |
| Mucoepidermoid carcinoma (n=23) | 0.78 | 0.87 | 1.00 |
| Acinar cell carcinoma (n=18) | 0.89 | 1.00 | 1.00 |
| Adenocarcinoma (n=15) | 0.53 | 0.73 | 0.80 |
| Malignant myoepithelioma (n=7) | 0.14 | 0.57 | 0.57 |
| Infiltrating duct carcinoma (n=5) | 0.20 | 0.40 | 0.40 |
| Adenosquamous carcinoma (n=4) | 0.75 | 0.75 | 1.00 |
| Epithelial-myoepithelial carcinoma (n=3) | 0.67 | 0.67 | 0.67 |
| Malignant melanoma (n=2) | 1.00 | 1.00 | 1.00 |
| Basal cell adenocarcinoma (n=2) | 0.00 | 0.50 | 0.50 |
| Carcinoid tumor (n=2) | 1.00 | 1.00 | 1.00 |

| Atlas2-S | Prec | Rec | F1 |
|---|---|---|---|
| Squamous cell carcinoma (n=29) | 0.77 | 0.79 | 0.78 |
| Adenoid cystic carcinoma (n=27) | 0.66 | 0.78 | 0.71 |
| Mucoepidermoid carcinoma (n=23) | 0.73 | 0.70 | 0.71 |
| Acinar cell carcinoma (n=18) | 0.94 | 0.83 | 0.88 |
| Adenocarcinoma (n=15) | 0.57 | 0.27 | 0.36 |

| Atlas2-B | Prec | Rec | F1 |
|---|---|---|---|
| Squamous cell carcinoma (n=29) | 0.71 | 0.83 | 0.76 |
| Adenoid cystic carcinoma (n=27) | 0.70 | 0.70 | 0.70 |
| Mucoepidermoid carcinoma (n=23) | 0.81 | 0.74 | 0.77 |
| Acinar cell carcinoma (n=18) | 0.89 | 0.89 | 0.89 |
| Adenocarcinoma (n=15) | 0.62 | 0.53 | 0.57 |

| UNI2h | Prec | Rec | F1 |
|---|---|---|---|
| Squamous cell carcinoma (n=29) | 0.77 | 0.79 | 0.78 |
| Adenoid cystic carcinoma (n=27) | 0.70 | 0.78 | 0.74 |
| Mucoepidermoid carcinoma (n=23) | 0.64 | 0.61 | 0.62 |
| Acinar cell carcinoma (n=18) | 1.00 | 0.94 | 0.97 |
| Adenocarcinoma (n=15) | 0.60 | 0.40 | 0.48 |

| Atlas2 | Prec | Rec | F1 |
|---|---|---|---|
| Squamous cell carcinoma (n=29) | 0.75 | 0.83 | 0.79 |
| Adenoid cystic carcinoma (n=27) | 0.76 | 0.70 | 0.73 |
| Mucoepidermoid carcinoma (n=23) | 0.78 | 0.78 | 0.78 |
| Acinar cell carcinoma (n=18) | 0.94 | 0.89 | 0.91 |
| Adenocarcinoma (n=15) | 0.73 | 0.53 | 0.62 |

**Figure 3. Per-class retrieval on the pathologist-selected query cohort.** (a) Comparison of the foundation models' recall performance per ICD-O class at Top-1, Top-3 and Top-5. (b) Precision, recall and F1 of the Top-1 retrieval for the five classes with at least ten cases. (c) Top-1/3/5 recall, averaged across the four foundation models, compared with chance-level retrieval, for the seven classes with at least five cases. A dot is filled only if every one of the four models individually exceeded chance. Prec = Precision, Rec = Recall.

Next, we evaluated the discriminatory power of our search engine. AUC was calculated based on the Top-5 retrieved cases (Fig. 4a). Atlas2 demonstrated a significantly higher AUC (0.754) than its distilled counterparts Atlas2-B (0.679, p = 0.020) and Atlas2-S (0.653, p = 0.003), but not UNI2h (0.697, p = 0.196).

When analyzing the Chamfer score distributions between correctly and incorrectly retrieved cases, all models showed significant discriminatory power, with larger models outperforming smaller variants (Fig. 4b). Absolute score levels decreased with increasing representation dimensionality within the Atlas family: Atlas2-S produced the highest scores (mean: 1.49) but the weakest retrieval and Atlas2 the lowest scores (mean: 1.34) and the strongest retrieval.

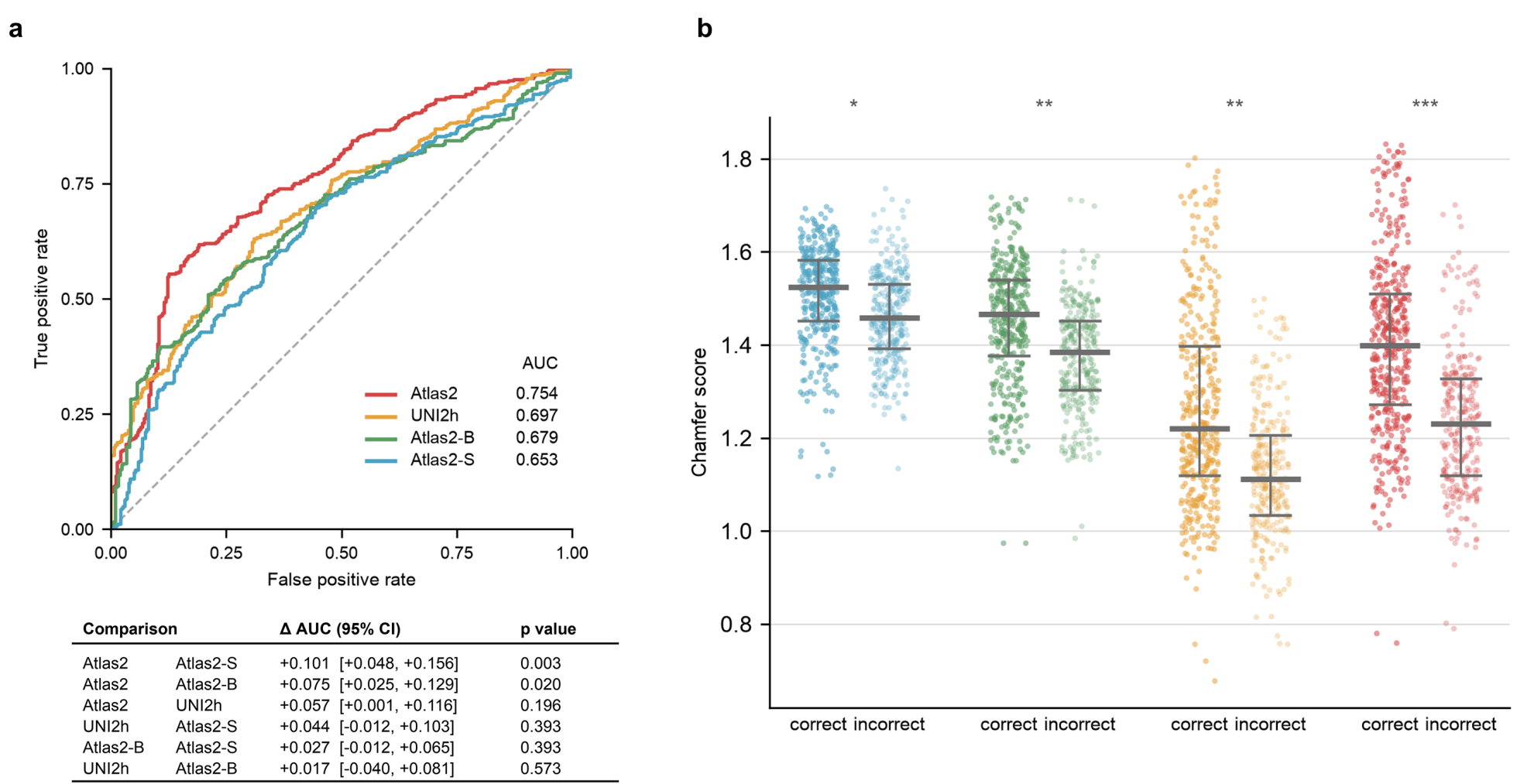

| Comparison | | Δ AUC (95% CI) | p value |
|---|---|---|---|
| Atlas2 | Atlas2-S | +0.101 [+0.048, +0.156] | 0.003 |
| Atlas2 | Atlas2-B | +0.075 [+0.025, +0.129] | 0.020 |
| Atlas2 | UNI2h | +0.057 [+0.001, +0.116] | 0.196 |
| UNI2h | Atlas2-S | +0.044 [-0.012, +0.103] | 0.393 |
| Atlas2-B | Atlas2-S | +0.027 [-0.012, +0.065] | 0.393 |
| UNI2h | Atlas2-B | +0.017 [-0.040, +0.081] | 0.573 |

**Figure 4. Discriminatory power and Chamfer score distributions.** (a) Area under the receiver operating characteristic curves and pairwise between-model comparisons with 95% confidence intervals and p-values. (b) Distribution of chamfer scores for correctly and incorrectly retrieved cases. Horizontal bars indicate the median and interquartile range ($p < 0.001$, $**p < 10^{-8}$, $***p < 10^{-15}$).

## Discussion

Content-based retrieval with off-the-shelf pathology FMs identified the correct diagnosis among the top five candidates in up to 81.3% of cases, without any task-specific training or manual annotation. This was shown on salivary gland tumors, a rare and diagnostically challenging group, demonstrating that current FMs allow pathologists to find rare entities in institutional archives.

While Top-1 recall varied across FMs (50–63%), it narrowed at Top-5 (73–81%), without any statistically significant difference between models at any depth. Macro-averaged F1 showed a similarly modest spread, from 68.8% for the smallest FM, Atlas2-S, to 76.6% for the largest, Atlas2. Chamfer scores between correct and incorrect retrievals overlapped considerably reflecting the overlapping morphological features characteristic of salivary gland tumors. Atlas2 achieved significantly better ranking performance than its distilled counterparts, Atlas2-B and Atlas2-S, but showed no significant difference relative to UNI2h. Discriminatory power rose with the dimensionality of the representations as absolute similarity fell, suggesting that larger representations encode finer tissue detail that better separates correct from incorrect matches. The effective granularity of retrieval is limited by the granularity of the diagnostic labels as reflected by the comparatively poor performance for adenocarcinoma and the perfect retrieval of two-case entities such as malignant melanoma and carcinoid tumor.

Searching histopathological images is computationally and biologically challenging given the size of WSIs, long tail of rare entities, inter-patient heterogeneity, and processing variability. SMILY was one of the first approaches, comparing representations extracted with a convolutional neural network (10). Yottixel and SISH then introduced strategies to reduce a WSI to a small set of representative patches, thereby facilitating search (11, 12). Our design responds to both constraints with a mixed strategy: on the query side, pathologists choose a query region, which is only a fraction of a WSI limiting computational work. As the composition of a WSI is primarily a property of grossing, a small set of representative regions carries most of the diagnostically relevant signal (26). Our split in homogeneous vs. heterogeneous queries still leaves room for intra-tumor heterogeneity (27). On the archive side, we index all artifact-free patches, which evades expert annotation work and avoids the risk of discarding diagnostically decisive regions. Through product quantization and inverted-file structure, the indices can be queried in standard RAM configurations (25). Our modified chamfer similarity is a middle ground between patch-to-patch and WSI-to-WSI retrieval, balancing speed and accuracy.

Prior AI work on salivary gland histopathology has focused on classification tasks, performing well at benign/malignant discrimination but degrading on histologic subtyping (28, 29). Fitting the correct decision boundaries is limited by the number of examples per class, which becomes increasingly difficult in the long tail of a taxonomy (30) and typically requires manual annotation beyond the routinely encoded ICD-O code (31). While AI decision aids improve accuracy when their output is correct, they reduce it when it is wrong (32), a problem particularly pronounced for ambiguous entities like salivary gland neoplasms where a Dutch series recently reported 8.3% discordance on expert review (33). By presenting the pathologist with a shortlist rather than a single label, our approach may be less prone to this confirmatory anchoring and can guide more effective ancillary testing. A content-based retrieval approach in oral pathology was recently shown to increase diagnostic accuracy, with the largest gains in underrepresented tumor categories (34).

Despite evaluating a broad cohort of a rare tumor group, we could not represent its full morphological spectrum. All cases originate from a single institution, digitized on the same scanners, which limits external validity. Since we use FMs pretrained on multi-institutional data, we expect the approach to transfer well to other institutions and entities. All diagnoses were made by board-certified pathologists, yet they were rendered over a period of years, not reflecting the recently updated ICD-O-4 classification (35). Algorithmically, product quantization and approximate nearest-neighbor search do not guarantee retrieval of the mathematically most similar case, yet this aligns with our approach of providing pathologist with a set of most similar cases. While we regard pathologist-selected query regions as the best way to use our search engine, an automatic WSI-to-WSI mode is available through our ABMIL model. Its lower performance compared with pathologist-selected queries indicates that it does not identify the most relevant query regions, most likely due to limited training data.

In summary, we show that pretrained FMs, without further training, can extend an individual pathologist's frame of reference to that of the department. Future work will test whether the tool measurably improves pathologists' performance and how the approach scales to larger cohorts.

**Acknowledgments**

The authors gratefully acknowledge LMU Klinikum for providing computing resources on their Clinical Open Research Engine (CORE), and Aignostics GmbH for providing access to Atlas2 foundation models under a research license.

**Author Contributions**

M.J.R and F.K. conceived and designed the study. M.J.R. developed the software and performed the formal analysis. N.K. and P.K. performed data curation. F.K. supervised the project. M.J.R. drafted the original manuscript. All authors contributed to writing, review, and revision, and read and approved the final version of the paper.

**Data Availability**

Anonymized data and the code for the retrieval pipeline are available upon reasonable request. Requests may be sent to the corresponding author.

**Declaration of Competing Interests**

F.K. is a co-founder, lead medical advisor and board member of Aignostics. The other authors declare no conflict of interest.

**Ethics Approval and Consent to Participate**

The study was approved by the Ethics Committee of the Medical Faculty of the Ludwig-Maximilians-Universität München (No. 26-0231). Written informed consent was waived due to the retrospective study design and the use of deidentified data.

**Funding**

This project has received support from the Bavarian Cancer Research Center (BZKF).

## References

1. Rao A, Barkley D, França GS, Yanai I. Exploring tissue architecture using spatial transcriptomics. Nature. 2021;596(7871):211-20.

2. Jaarsma T, Jarodzka H, Nap M, van Merriënboer JJG, Boshuizen HPA. Expertise in clinical pathology: combining the visual and cognitive perspective. Adv in Health Sci Educ. 2015;20(4):1089-106.

3. Tizhoosh HR, Pantanowitz L. On image search in histopathology. Journal of Pathology Informatics. 2024;15:100375.

4. van der Laak J, Litjens G, Ciompi F. Deep learning in histopathology: the path to the clinic. Nat Med. 2021;27(5):775-84.

5. LeCun Y, Bengio Y, Hinton G. Deep learning. Nature. 2015;521(7553):436-44.

6. Campanella G, Hanna MG, Geneslaw L, Miraflor A, Werneck Krauss Silva V, Busam KJ, Brogi E, Reuter VE, Klimstra DS, Fuchs TJ. Clinical-grade computational pathology using weakly supervised deep learning on whole slide images. Nat Med. 2019;25(8):1301-9.

7. Chen RJ, Ding T, Lu MY, Williamson DFK, Jaume G, Song AH, Chen B, Zhang A, Shao D, Shaban M, Williams M, Oldenburg L, Weishaupt LL, Wang JJ, Vaidya A, Le LP, Gerber G, Sahai S, Williams W, Mahmood F. Towards a general-purpose foundation model for computational pathology. Nat Med. 2024;30(3):850-62.

8. Alber M, Milbich T, Carpen-Amarie A, Tietz S, Dippel J, Muttenthaler L, Cancer BP, Benetti A, Korfiatis P, Eulig E, Lüscher J, Wu J, Hashimi SA, Dernbach G, Schallenberg S, Shah N, Krügener M, Jammoria A, Matras J, Duffy P, Redlon M, Jurmeister P, Horst D, Ruff L, Müller K-R, Klauschen F, Norgan A. Atlas 2 — Foundation models for clinical deployment. arXiv [Internet]. 2026 [cited arXiv:2601.05148. Available from: https://arxiv.org/abs/2601.05148.

9. Ghassemi M, Oakden-Rayner L, Beam AL. The false hope of current approaches to explainable artificial intelligence in health care. The Lancet Digital Health. 2021;3(11):e745-e50.

10. Hegde N, Hipp JD, Liu Y, Emmert-Buck M, Reif E, Smilkov D, Terry M, Cai CJ, Amin MB, Mermel CH, Nelson PQ, Peng LH, Corrado GS, Stumpe MC. Similar image search for histopathology: SMILY. npj Digit Med. 2019;2(1):56.

11. Kalra S, Tizhoosh HR, Choi C, Shah S, Diamandis P, Campbell CJV, Pantanowitz L. Yottixel – An Image Search Engine for Large Archives of Histopathology Whole Slide Images. Medical Image Analysis. 2020;65:101757.

12. Chen C, Lu MY, Williamson DFK, Chen TY, Schaumberg AJ, Mahmood F. Fast and scalable search of whole-slide images via self-supervised deep learning. Nature Biomedical Engineering. 2022;6(12):1420-34.

13. Jansen L, Mayer M, Stang A, Kajueter H, Oesterling F, Szentkiralyi A, Kourou S, Wolber P, van de Loo J, Hieggelke L, Quaas A, Klussmann JP, Nachtsheim L. Incidence and survival of primary major salivary gland carcinoma in Germany over the last decade: A nationwide population-based study. European Journal of Cancer. 2026;234:116196.

14. Skálová A, Hyrcza MD, Leivo I. Update from the 5th Edition of the World Health Organization Classification of Head and Neck Tumors: Salivary Glands. Head and Neck Pathol. 2022;16(1):40-53.

15. Sarangi J, Kakkar A, Roy D, Mishra D, Thakar A, Deo SVS, Sharma A, Bhasker S. Metastases to the Parotid Gland: Study from a Tertiary Care Centre. Head and Neck Pathol. 2022;16(4):1034-42.

16. Higgins KE, Cipriani NA. Practical immunohistochemistry in the classification of salivary gland neoplasms. Seminars in Diagnostic Pathology. 2022;39(1):17-28.

17. Jurmeister P, Leitheiser M, Arnold A, Capilla EP, Mochmann LH, Zhdanovic Y, Schleich K, Jung N, Chimal EC, Jung A, Kumbrink J, Harter P, Prenißl N, Elezkurtaj S, Brcic L, Deigendesch N, Frank S, Hench J, Försch S, Breimer G, van Engen van Grunsven I, Lassche G, van Herpen C, Zhou F, Snuderl M, Agaimy A, Müller K-R, von Deimling A, Capper D, Klauschen F, Ihrler S. DNA Methylation Profiling of Salivary Gland Tumors Supports and Expands Conventional Classification. Modern Pathology. 2024;37(12):100625.

18. Weng Z, Seper A, Pryalukhin A, Mairinger F, Wickenhauser C, Bauer M, Glamann L, Bläker H, Lingscheidt T, Hulla W, Jonigk D, Schallenberg S, Bychkov A, Fukuoka J, Braun M, Schömig-Markiefka B, Klein S, Thiel A, Bozek K, Netto GJ, Quaas A, Büttner R, Tolkach Y. GrandQC: A comprehensive solution to quality control problem in digital pathology. Nat Commun. 2024;15(1):10685.

19. Hörst F, Schaheer SH, Baldini G, Bahnsen FH, Egger J, Kleesiek J. Accelerating Artificial Intelligence-based Whole Slide Image Analysis with an Optimized Preprocessing Pipeline. Bildverarbeitung für die Medizin 2024 (Informatik aktuell): Springer Fachmedien Wiesbaden; 2024. p. 356-61.

20. Ding T, Wagner SJ, Song AH, Chen RJ, Lu MY, Zhang A, Vaidya AJ, Jaume G, Shaban M, Kim A, Williamson DFK, Robertson H, Chen B, Almagro-Pérez C, Doucet P, Sahai S, Chen C, Chen CS, Komura D, Kawabe A, Ochi M, Sato S, Yokose T, Miyagi Y, Ishikawa S, Gerber G, Peng T, Le LP, Mahmood F. A multimodal whole-slide foundation model for pathology. Nat Med. 2025;31(11):3749-61.

21. Oquab M, Darcet T, Moutakanni T, Vo H, Szafraniec M, Khalidov V, Fernandez P, Haziza D, Massa F, El-Nouby A, Assran M, Ballas N, Galuba W, Howes R, Huang P-Y, Li S-W, Misra I, Rabbat M, Sharma V, Synnaeve G, Xu H, Jegou H, Mairal J, Labatut P, Joulin A, Bojanowski P. DINOv2: Learning Robust Visual Features without Supervision. arXiv [Internet]. 2023 [cited arXiv:2304.07193. Available from: https://arxiv.org/abs/2304.07193.

22. Darcet T, Oquab M, Mairal J, Bojanowski P. Vision Transformers Need Registers. arXiv [Internet]. 2023 [cited arXiv:2309.16588. Available from: https://arxiv.org/abs/2309.16588.

23. Ilse M, Tomczak JM, Welling M, editors. Attention-based Deep Multiple Instance Learning. Proceedings of the 35th International Conference on Machine Learning (ICML), PMLR 80; 2018. arXiv:1802.04712: PMLR.

24. Lu MY, Williamson DFK, Chen TY, Chen RJ, Barbieri M, Mahmood F. Data-efficient and weakly supervised computational pathology on whole-slide images. Nature Biomedical Engineering. 2021;5(6):555-70.

25. Douze M, Guzhva A, Deng C, Johnson J, Szilvasy G, Mazaré P-E, Lomeli M, Hosseini L, Jégou H. The Faiss Library. IEEE Trans Big Data. 2026;12(2):346-61.

26. Alsaafin A, Nejat P, Shafique A, Khan J, Alfasly S, Alabtah G, Tizhoosh HR. Sequential Patching Lattice for Image Classification and Enquiry. The American Journal of Pathology. 2024;194(10):1898-912.

27. Song AH, Chen RJ, Ding T, Williamson DFK, Jaume G, Mahmood F, editors. Morphological Prototyping for Unsupervised Slide Representation Learning in Computational Pathology. 2024 IEEE/CVF Conference on Computer Vision and Pattern Recognition (CVPR); 2024.

28. Weber A, Schuster D, Heyer J, Becker C, Burkhardt V, Werner M, Spörlein A, Bronsert P, Schulz T. Deep learning-based histopathological classification and subclassification of benign and malignant salivary gland tumors. Eur Arch Otorhinolaryngol. 2026;283(6):3905-15.

29. Alsanie I, Shephard A, Azarmehr N, Vargas P, Pring M, Rajpoot NM, Khurram SA. Exploring the feasibility of AI-based analysis of histopathological variability in salivary gland tumours. Sci Rep. 2025;15(1):29171.

30. Skálová A, Laco J, Thompson LDR, Bradová M, Vander Poorten V, Araújo ALD, Stenman G, Leivo I, Agaimy A, Ferlito A. Recent developments in salivary gland pathology after the WHO 2024 classification: new developments in existing entities and evolving new entities. Virchows Arch. 2026;488(5):1005-20.

31. Li M, Shen Z-l, Xian H-c, Zheng Z-j, Yu Z-w, Liang X-h, Gao R, Tang Y-l, Zhang Z. A Recognition System for Diagnosing Salivary Gland Neoplasms Based on Vision Transformer. The American Journal of Pathology. 2025;195(2):221-31.

32. Kiani A, Uyumazturk B, Rajpurkar P, Wang A, Gao R, Jones E, Yu Y, Langlotz CP, Ball RL, Montine TJ, Martin BA, Berry GJ, Ozawa MG, Hazard FK, Brown RA, Chen SB, Wood M, Allard LS, Ylagan L, Ng AY, Shen J. Impact of a deep learning assistant on the histopathologic classification of liver cancer. npj Digit Med. 2020;3(1):23.

33. Reerds STH, Uijen MJM, Van Engen-Van Grunsven ACH, Marres HAM, van Herpen CML, Honings J. Results of histopathological revisions of major salivary gland neoplasms in routine clinical practice. J Clin Pathol. 2023;76(6):374-8.

34. Herdiantoputri RR, Komura D, Ochi M, Fukawa Y, Oba K, Tsuchiya M, Kikuchi Y, Matsuyama Y, Ushiku T, Ikeda T, Ishikawa S. Preclinical Evaluation of an Interactive Image Search System of Oral Pathology. J Dent Res. 2025;104(11):1192-201.

35. Znaor A, Rous B, Levy GG, Watanabe R, Jakob R, Krpelanova E, Wijesinghe HD, Puspanathan P, Cree IA, Bray F, Lokuhetty D. The International Classification of Diseases for Oncology, 4th Edition (ICD-O-4): An overview. Cancer Epidemiology. 2026;101:102989.